\documentclass{article}

\usepackage[preprint]{neurips_2024}

\usepackage[utf8]{inputenc} % allow utf-8 input
\usepackage[T1]{fontenc}    % use 8-bit T1 fonts
\usepackage{hyperref}       % hyperlinks
\usepackage{url}            % simple URL typesetting
\usepackage{booktabs}       % professional-quality tables
\usepackage{amsfonts}       % blackboard math symbols
\usepackage{nicefrac}       % compact symbols for 1/2, etc.
\usepackage{microtype}      % microtypography
\usepackage{xcolor}         % colors
\usepackage{graphicx}
\usepackage{amsmath} 
\usepackage{tikz}
\usetikzlibrary{arrows.meta}
\usepackage{multirow}
\usepackage{makecell}
\usepackage{algorithm}
\usepackage{algorithmic}

\title{Online Decision MetaMorphFormer: A Casual Transformer-Based Reinforcement Learning Framework of Universal Embodied Intelligence}

\author{%
  Luo Ji\textsuperscript{1} \thanks{Corresponding author: \texttt{jiluo.lj@alibaba-inc.com}}, Runji Lin\textsuperscript{1} \\ 
  \textsuperscript{1} DAMO Academy, Alibaba Group\\
  \texttt{\href{https://rlodm.github.io/odm/}{https://rlodm.github.io/odm/}} \\
}
\begin{document}
\maketitle

\begin{abstract}
Interactive artificial intelligence in the motion control field is an interesting topic, especially when universal knowledge is adaptive to multiple tasks and universal environments. Despite there being increasing efforts in the field of Reinforcement Learning (RL) with the aid of transformers, most of them might be limited by the offline training pipeline, which prohibits exploration and generalization abilities. To address this limitation, we propose the framework of Online Decision MetaMorphFormer (ODM) which aims to achieve self-awareness, environment recognition, and action planning through a unified model architecture. Motivated by cognitive and behavioral psychology, an ODM agent is able to learn from others, recognize the world, and practice itself based on its own experience.  ODM can also be applied to any arbitrary agent with a multi-joint body, located in different environments, and trained with different types of tasks using large-scale pre-trained datasets. Through the use of pre-trained datasets, ODM can quickly warm up and learn the necessary knowledge to perform the desired task, while the target environment continues to reinforce the universal policy. Extensive online experiments as well as few-shot and zero-shot environmental tests are used to verify ODM's performance and generalization ability. The results of our study contribute to the study of general artificial intelligence in embodied and cognitive fields. Code, results, and video examples can be found on the website \url{https://rlodm.github.io/odm/}.
\end{abstract}

\section{Introduction}

Research of embodied intelligence focus on the learning of control policy given the agent with some morphology (joints, limbs, motion capabilities), while it has always been a topic whether the control policy should be more general or specific. As the improvement of large-scale data technology and cloud computing ability, the idea of artificial general intelligence (AGI) has received substantial interest \cite{reed2022GATO}. Accordingly, a natural motivation is to develop a universal control policy for different morphological agents and easy adaptive to different scenes. It is argued that such a smart agent could be able to identify its 'active self' by recognizing the egocentric, proprioceptive perception, react with exteroceptive observations and have the perception of world forward model \cite{hoffmann2012emobodiment}. However, there is seldom such machine learning framework by so far although some previous studies have similar attempts in one or several aspects.
% recognize embodiment cognition and
 
%Challenge of new environments. adaptation.
%motion control.
%Egocentric Perception \citep{Brooks1999CogProj}
%proprioception
%Active self
%Peripersonal Space (PPS)
%Intelligence requires a body. body schema, forward model.
%human cognitive science: learning from pioneers, predict the world, and evolve with experience.

%for a novel and effective model architecture called transformer, which tries to model a large-scale sequence problem and train the model in a self-regressive way. Correspondingly, transformer-based RL \citep{chen2021DT, lee2022multigameDT, janner2021TT, zheng2022ODT, xu2022PromptDT}. However, most of these studies are offline RL which build a big sequence model problem and tries to learning from large-scale offline dataset. More emphasize on agent morphology. For example, \citet{kurin2020amorpheus} propose Amorpheus which encode each body joint as a transformer but disregard the connection information between joints. On the other hand, MetaMorph \citep{gupta2022MetaMorph} innovatively use transformer to encode the body morphology as a sequence of joint properties, instead of time sequence dependency. 
%two-directional 

Reinforcement Learning(RL) learns the policy interactively based on the environment feedback therefore could be viewed as a general solution for our embodied control problem. Conventional RL could solve the single-task problem in an online paradigm, but is relatively difficult to implement and slow in practice, and lack of generalization and adaptation ability. Offline RL facilitates the implementation but in cost of performance degradation. Inspired by recent progress of large model on language and vision fields, transformer-based RL \cite{reed2022GATO, chen2021DT, lee2022multigameDT, janner2021TT, zheng2022ODT, xu2022PromptDT} has been proposed by transforming RL trajectories as a large time sequence model and train it in the auto-regressive manner. However, most of these studies belong to offline RL which build a big sequence model problem and tries to learning from large-scale offline dataset. Such methods provide an effective approach to train a generalist agent for different tasks and environments, but usually have worse performance than classic RL, and fail to capture the morphology information. In contrast, MetaMorph \cite{gupta2022MetaMorph} chooses to encode on agent's body morphology and performs online learning, therefore has good performance but lack of time-dependency consideration.

%To have a better solution of embodied intelligence, we are motivated from behavioral psychology in which agent improve its skill by actual practice, learning from others (teachers, peers or even someone with worse skills), or makes decision based on the perception of 'the world model' \cite{ha2018RecurrentWorldModel, wu2022DayDreamer}.  It is reasonable to believe that an embodied intelligence agent should have the above three learning paradigm simultaneously. We propose such a methodology by designing a morphology-time transformer-based RL architecture which is compatible with both offline and online learning. Offline training is conducted on multi-task datasets which considers both learning from other agents and speculate the future system states. The online training allows the agent to improve its policy in an on-policy way given a single task.

%\vspace{-3mm}
\begin{figure*}[t!]

\begin{center}
%\framebox[4.0in]{$\;$}s
\includegraphics[width=1.0\linewidth]{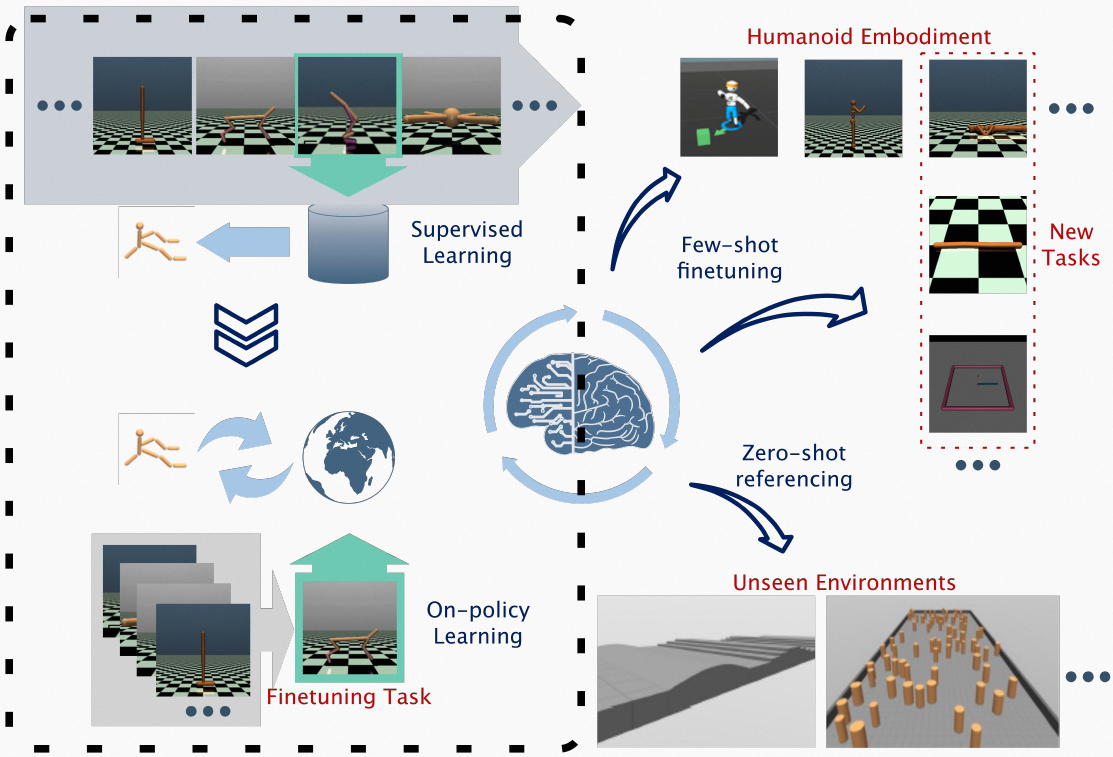} %,height=6cm
%\includegraphics[width=1.0\linewidth]{application_pipeline_p.pdf}
%\fbox{\rule[-.5cm]{0cm}{4cm}  \rule[-.5cm]{4cm}{0cm}}
\end{center}
\caption{Application pipeline of ODM.} \label{pipeline}
%\vspace{-2mm}
\end{figure*}

In this work, we propose a framework called \textbf{O}nline \textbf{D}ecision \textbf{M}etamorphformer (ODM), which aims to study the general knowledge of embodied control across different body shapes, environments and tasks, as indicated in Figure \ref{pipeline}. The idea is motivated from the behavioral psychology in which agents improve their skill by actual practice, learning from others (teachers, peers or even someone with worse skills), or makes decision based on the perception of 'the world model' \cite{ha2018RecurrentWorldModel, wu2022DayDreamer}. Our ODM is built upon a morphology-time transformer-based RL architecture and is compatible with both offline and online learning. The model contains the universal backbone and the task-specific modules. The task-specific modules capture the potential difference in agent body shapes, and the morphological difference is enhanced by a prompt based on characteristic of body shapes. We first pretrain this model with a curriculum learning, by learning demonstrations from the easiest to the hardest task, from the expert to beginners. The environment model prediction is added as an auxiliary loss. The same architecture can then be finetuned online given a specific task. During the test, we are able to test ODM with all training environments, transfer the policy to different body shapes, adaptive to unseen environments and accommodate with new types of tasks (e.g. from locomotion to reaching, target capturing or escaping from obstacles.).
%code is \url{https://anonymous.4open.science/r/OnlineDecisionMetaMorphFormer-3F7F}

% The morphology-aware auto-encoder is employed to encode states and actions and map to the same latent space. A casual transformer is then applied on this sequence problem to learn the general control knowledge, but with a morphological prompt to capture each body shape's characteristic.
%  with the very last action trained by on-policy gradient method, and another value prediction head is added and learned by bootstrap method

%imilar with Trajectory Transformer, we build a unified sequence model to model the episode and conduct actor-critic RL, imitation learning, model-based RL at the same time. However, we in the basis of continuous space, instead of discrete space to better take advantage of the dense information in motion control tasks.

Main contributions of this paper include:
\begin{itemize}
    \item We design a unified model architecture to encode time and morphology dependency simultaneously which bridges sequential decision making with embodiment intelligence.
    \item We propose a training paradigm which mimic the process of natural intelligence emerging, including learning from others, boost with practices, and recognize the world.
    \item We train and test our framework with agent in eight different body shapes, different environment terrain and different task types. These comprehensive analysis verifies the general knowledge of motion control learned by us.
\end{itemize}

%Unity \citep{juliani2020unity}
%unimal \citep{gupta2021unimal}

%MetaMorph \citep{gupta2022MetaMorph}
%decision transformer \citep{chen2021DT}
%multi-game decision transformer \citep{lee2022multigameDT}
%trajectory transformer \citep{janner2021TT}
%online decision transformer \citep{zheng2022ODT}
%multi-agent decision transformer \citep{meng2021MADT}
%Prompt-DT \citep{xu2022PromptDT}

%Amorpheus \citep{kurin2020amorpheus}
%PPO \citep{schulman2017ppo}
%World Model \citep{ha2018RecurrentWorldModel}
%DayDreamer \citep{wu2022DayDreamer}

The rest of the paper is organized as follows. The connection with previous works is first discussed in Related Work. We then introduce the preliminaries and the problem formulation. Our methodology and corresponding algorithms are stated in the Method. We summarize our results and compare with previous baselines in Experiment. We discuss the shortcomings and potential improvements in Discussion and finally concludes the paper.

% Section \ref{sec:related_work}
% \ref{sec:preliminaries}
% \ref{sec:method}
% \ref{sec:experiment}
% Section \ref{sec:discussion}
% in Section \ref{sec:conclusion}

\section{Related Work}
\label{sec:related_work}

\noindent \textbf{Classic RL}: Among conventional RL methods, on-policy RL such as Proximal Policy Optimization (PPO) \cite{schulman2017ppo} is able to learn the policy and therefore has a good adaptive ability to environment, but is slow to convergence and might have large trajectory variations. Off-policy RL such as DQN \cite{mnih2015DQN} improves the sampling efficiency but still requires the data buffer updated dynamically. In contrast, offline RL \cite{Fujimoto2019BCQ, Kumar2020CQL, kostrikov2021IQL} can solve the problem similar to supervised learning, but might have degraded performance because of the distribution shift between offline dataset and online environment. In our work, we aim to reach state-of-the-art performance for different embodied control tasks, therefore a model architecture compatible with on-policy Rl is proposed.

\noindent \textbf{Transformer-based RL}: Among these efforts, Decision Transformer (DT) \cite{chen2021DT} and Multi-game Decision Transformer \cite{lee2022multigameDT} embodied the continuous state and action directly and use a GPT-like casual transformer to solve the policy offline. Their action decision is conditioned on Return-to-Go (RTG), either arbitrarily set or estimated by the model, since RTG is unknown during inference. Instead, Trajectory Transformer (TT) \cite{janner2021TT} discards RTG in the sequential modeling to avoid that approximation.PromptDT \cite{xu2022PromptDT} adds the task difference consideration into the model by using demonstrated trajectory as prompts. ODT \cite{zheng2022ODT} first attempts to solve transformer-based RL in an online manner but mainly focus on supervised on actions instead of maximizing rewards. In our work, we propose a similar model architecture that is able to conduct both offline learning and on-policy, actor-critic learning. Online learning employs PPO as the detailed policy update tool with the objective as reward maximization.

%  TT has a similar model architecture and training paradigm but tokenizing the state and action discretely. 

\noindent \textbf{Morphology-based RL}: There are some other studies that focus on the agent's morphology information, including GNN-based RL which models agent joints as a kinematics tree graph \cite{huang2020smp}, Amorpheus \cite{kurin2021amorpheus} which encodes a policy modular for each body joint, and MetaMorph \cite{gupta2022MetaMorph} which intuitively use a transformer to encode the body morphology as a sequence of joint properties and train it by PPO. In our work, we have the morphology-aware encoder which is similar with MetaMorph and has the same PPO update rule. However, compared with MetaMorph, we encode the morphology on not only the state but also historical actions, and consider the historical contextual consideration.

\section{Preliminaries and Problem Setup}
\label{sec:preliminaries}

%\subsection{Reinforcement Learning}
%\label{sec:rl}

%  generally studies on \textit{Markov Decision Process} (MDP).
\subsection{Reinforcement Learning} 

We formulate a typical sequential decision-making problem, in which on each time step $t$, an embodied agent conceives a state $s_t \in \mathcal{R}^{n_s}$, performs an action $a_t \in \mathcal{R}^{n_a}$, and receives a scalar reward $r_t \in \mathcal{R}^1$. Given the current state, the agent generates an action from its policy $\pi(a_t | s_t)$ and push the environment stage forward. This interactive process yields the following episode sequence: 
\begin{equation}
    \tau_{0:T} = \{s_0, a_0, r_0, s_1, a_1, r_1, \cdots, s_T, a_T, r_T\}
\end{equation}
% (\mathcal{S, A, R, P})
in which $T$ means the episode ends or reaching the maximum time length. 

Reinforcement Learning (RL) is usually employed to solve such a problem by finding $\pi$ such that $\max_{\pi} \mathbb{E}_{\tau \sim \pi}(R)$ in which the episode return defined as
\begin{equation}
    R := \sum_{t=0}^{T} \gamma^t r_t \label{eq:return}
\end{equation}
with $\gamma \in [0, 1)$ as the discounted factor. 

%\vspace{-2mm}
\begin{table*}[h]
  \centering
  \caption{Comparison of Conventional RL and Our Notations}
  \label{tab:notation_comparison}
  %\begin{tiny}
  \begin{tabular}{@{}ccccc@{}}
    \hline
    \textbf{} & \textbf{State} & \textbf{Action}\\
    \hline
    Conventional RL & $s \in \mathcal{R}^{n_s}$ & $a \in \mathcal{R}^{n_a}$ \\ % [1pt]
    \hline
    \multirow{ 2}{*}{Our approach}  & $o^{pro}, \mathcal{M}_s^{pro} \in \mathbb{R}^{K \times n}, o^{ext} \in \mathbb{R}^x$ & \\
                            & $s = [o^{pro}, o^{ext}]$ & $a, \mathcal{M}_a \in \mathbb{R}^{K \times m}$ \\
    \hline
    Connections & $n_s = K*n - \sum \mathcal{M}_s + x$ & $n_a = K*m - \sum \mathcal{M}_a$ \\
    \hline
  \end{tabular}
  %\end{tiny}
\end{table*}
%\vspace{-2mm}

\subsection{Proximal Policy Optimization} 
\label{sec:ppo}

The classical Proximal Policy Optimization (PPO) methodology \cite{schulman2017ppo} inherits from the famous actor-critic framework in which a critic estimates the state value function $V(s)$, while an actor determines the policy. The critic loss is calculated from Bellman function by bootstrapping the state value function
\begin{equation}
\label{eq:loss_critic}
%\small
L^{\text{Critic}} = \mathbb{E}_{s \sim d^{\pi}}[r_t + \gamma (V_{\theta_{\text{old}}}(s_{t+1}) - V_{\theta}(s_{t}))^2] 
\notag
\end{equation}
On the other hand, generalized advantage estimation (GAE) \cite{schulman2016gae} is employed to help calculate the action advantage $A_t$  by traversing the episode backward
\begin{align}
    \hat{A}_t &= \delta_t + (\gamma \lambda) \delta_{t+1} + \cdots + (\gamma \lambda)^{T-t+1} \delta_{T-1} \notag \\
    \delta_t &= r_t + \gamma V(s_{t+1}) - V(s_t) \label{eq:gae}
\end{align}
the actor is then learned by maximizing the surrogate objective of $A_t$ according to the policy-gradient theorem
\begin{equation}
%\label{eq:loss_actor}
    L^{\text{Actor}} = \text{CLIP}(\text{E}_t \frac{\pi_{\theta_{\text{old}}}(a_t | s_t)}{\pi_{\theta}(a_t | s_t)} \hat{A_t} - \beta \text{KL} [\pi_{\theta_{\text{old}}}(\cdot | s_t), \pi_{\theta}(\cdot | s_t)]) \notag
\end{equation}
in which the $\text{CLIP}$ function means clipping the object by $[1 - \epsilon, 1 + \epsilon] \hat{A_t}$, and $\text{KL}$ denotes the famous K-L divergence. 
%Appendix \ref{sec:prelim_methods}.
A PPO policy update is then conducted by minimizing the objective upon each iteration:
\begin{equation}
\label{eq:loss_ppo}
L^{\text{PPO}} = -\eta_A L^{\text{Actor}} + \eta_C L^{\text{Critic}}
\end{equation}

%\subsection{Embodied Viewpoint of States and Actions}

\subsection{Problem Setup} 

Here we redefine the aforementioned conventional RL notations in a more `embodied style', although still generalized enough for any arbitrary agent with a multi-joint body. Inspired by the idea of \cite{gupta2022MetaMorph}, we differentiate the observation into the agent's proprioceptive observations, the agent's embodied joint-wise self-perception (e.g. angular, angular velocity of each joint), as well as the exteroceptive observation, which is the agent's global sensory (e.g. position, velocity). Given a $K$-joint agent, we denote the proprioceptive observation by $o^{pro} \in \mathbb{R}^{K \times n}$ in which each joint is embedded with $n$ dimension observations. The exteroceptive observation is $x$-dimensional which results in $o^{ext} \in \mathbb{R}^x$ and $s := [o^{pro}, o^{ext}]$.

Stepping forward from \cite{gupta2022MetaMorph}, we also define the action in the joint-dependent way; that is, assuming each joint has $m$ degree of freedom (DoF) of movements (e.g. torque), the action is reshaped as $a \in \mathbb{R}^{K \times m}$. To allow the room of different agent body shapes, we introduce binary masks which have the same shapes of $o^{pro}$ and $a$ and zero-pad the impossible observations or actions (e.g. DoF of a humanoid's forearm should be smaller than its upper-arm due to their physical connection). Table \ref{tab:notation_comparison} visualizes the comparison between the conventional RL notations and our embodied version notations.
%  and $K+1$ bodyparts (or limbs)

\subsection{Attention-based Encoding}
\label{sec:atten}

Given a stacked time sequence vector $x \in \mathbb{R}^{T \times e}$ with $T$ as the time length and $e$ as the embedding dimension, an time sequence encoder can be expressed as
%is formulated by applying Eq \ref{eq:attention} by feeding $Q, K, V$ all with $x$
\begin{equation}
\label{eq:time_encode}
\text{Enc}_T(x) = \text{Attention}(\text{Q=}x, \text{K=}x, \text{V=}x) \in \mathbb{R}^{T \times e}
\end{equation}
according to the self-attention mechanism \cite{vaswani2017attention} with Q, K, V denoting query, key and value. Analogously, given a stacked joint sequence vector $p \in \mathbb{R}^{K \times e}$ with $K$ as the number of joints, a morphology-aware encoder instead learns the latent representation on this joint sequence
\begin{equation}
\label{eq:morph_encode}
\text{Enc}_M(p) = \text{Attention}(\text{Q=}p, \text{K=}p, \text{V=}p) \in \mathbb{R}^{K \times e}
\end{equation}
By pre-tokenizing either $o^{pro}$ or $a$ into $p$, within the latent space with dimension $e$, their 'pose' latent variables can be encoded by $\text{Enc}_M$. 
For each form of encoder, timestep or joint position info is encoded by lookup embedding then adding to encoded vector. More details can be referred to the MetaMorph paper \cite{gupta2022MetaMorph}.
%This dot-product form of attention can be replaced by an alternative cosine similarity form.
%is applied in our self-attention block. We also apply another cosine similarity form of attention in our multi-head attention block.
%  with $Q, K, V$ using the same vector

% The attention mechanism \cite{vaswani2017attention} has the follow form:
%\begin{equation}
%\begin{split}
%    \text{Attention}(Q, K, V) = \text{softmax}(\frac{Q K^T}{\sqrt{d_k}}) V \label{eq:attention}
%\end{split}
%\end{equation}

%\vspace{-3mm}
\begin{figure*}[htbp!]
\begin{center}
%\framebox[4.0in]{$\;$}
\includegraphics[width=0.8\linewidth]{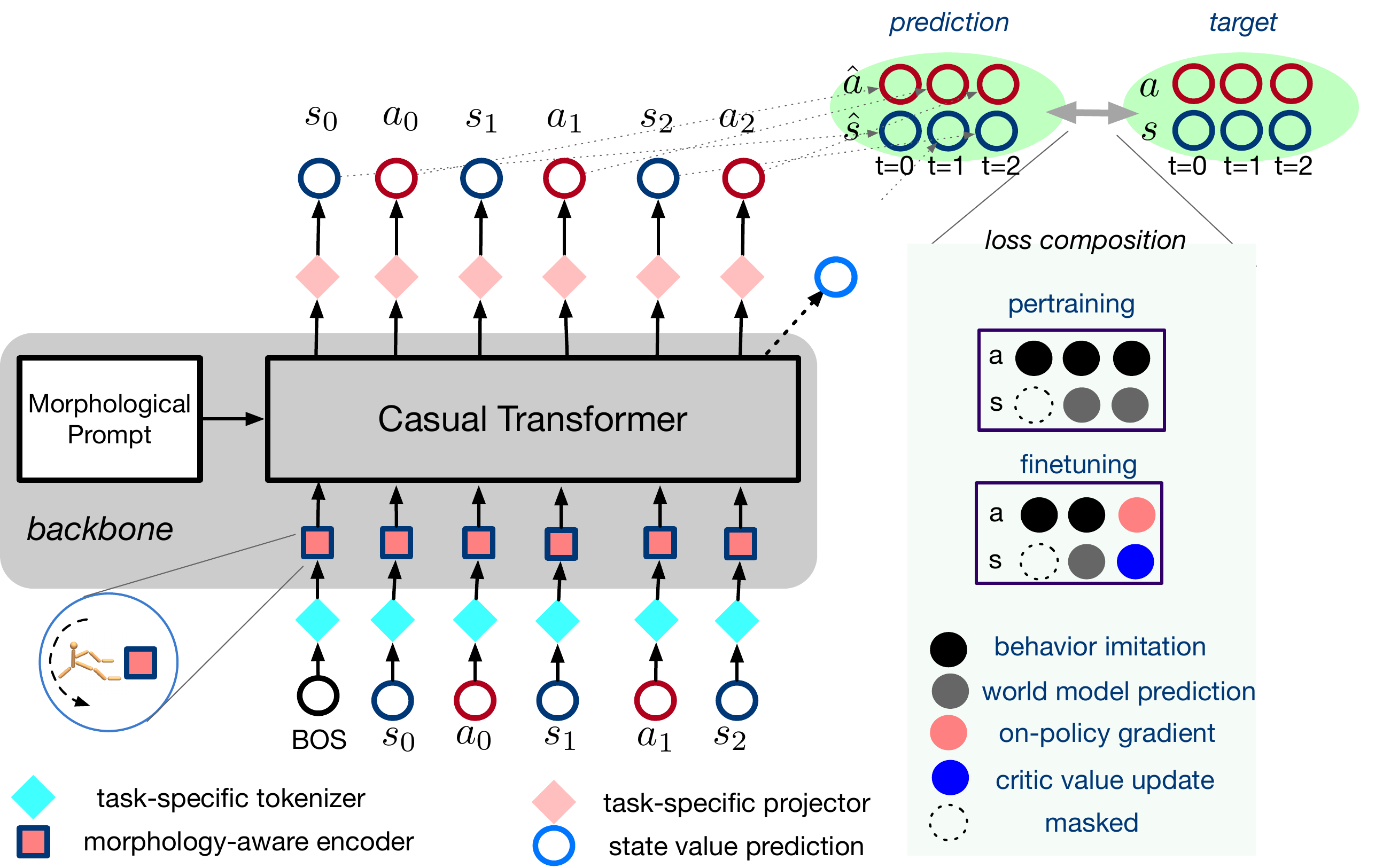} %,height=6cm
%\fbox{\rule[-.5cm]{0cm}{4cm}  \rule[-.5cm]{4cm}{0cm}}
\end{center}
\caption{Model Structure of ODM and Corresponding Training Paradigm.} \label{fig:model_structure}
\end{figure*}
%\vspace{-3mm}

\section{Methodology}
\label{sec:method}

%ODM is a unified framework for general embodied intelligence with morphological awareness, time sequence modeling and offline pertraining. Specifically, We design a special transformer model architecture with a morphology-dependent encoder and a decoder to model a trajactory sequence over time. For the training paradigm,  a general pre-training on diverse offline data and task-specific fine-tuning training paradigm are proposed to improve the generalization and sample efficiency.

%ODM is a unified framework for achieving general embodied intelligence, incorporating features such as morphological awareness, time sequence modeling, and offline pre-training. Specifically, we have developed a specialized transformer model architecture that includes a morphology-dependent encoder and decoder to model the trajectory sequence over time. In terms of training methodology, we have proposed a two-stage approach that involves general pre-training on diverse offline data followed by task-specific fine-tuning training to enhance both generalization and sample efficiency.

%ODM is a unified framework for achieving general embodied intelligence, incorporating features such as morphological awareness, time sequence modeling, and offline pre-training. Specifically, 

We have developed a specialized transformer model architecture that includes a morphology-dependent encoder and decoder to model the trajectory sequence over time. A two-stage training paradigm is proposed that involves general pre-training on diverse offline data followed by task-specific fine-tuning training.% to enhance both generalization and sample efficiency.

\subsection{Model Architecture}
\label{sec:model}

Our ODM model structure contains a unified backbone and task-specific modules. Within this paper's context, the task might be related to a different agent (potentially different body types), environment, and reward mechanisms. Since the original system variables might have different dimensions, task-specific modules map them into a uniform-dimensional latent space ($e$ in Eq. \ref{eq:morph_encode}) and reverse operations. The backbone has a two-directional, morphology-time transformer structure, including morphology-aware encoders and a casual transformer. Architecture details are specified in Fig \ref{fig:model_structure}. 
% Since our model deals with state and action variables in continuous spaces, 
% which deal with the time-dependent decision making process

%bMulti-Layer Perceptron
\noindent \textbf{Tokenizer} At each time $t$, observations and actions are first embedded into the latent space
\begin{align}
    o_t^e &= \text{Embed}_o(o^{pro}_t) \in \mathbb{R}^{K, e} \notag \\
    x_t^e &= \text{Embed}_x(o^{ext}_t) \in \mathbb{R}^{e} \notag \\
    a_t^e &= \text{Embed}_a(a_t) \in \mathbb{R}^{K, e}
    %s_t^e &= \text{MLP}_s([o_t^e, o^{ext}]) \in \mathbb{R}^{K, e} \\
\end{align}

\noindent \textbf{Morphology-aware Encoder} Corresponding pose embedding vectors are obtained by traversing the agent's kinematic tree and encoding the morphology by Eq. \ref{eq:morph_encode}:
\begin{align}
    o_t^p &= \text{Enc}_M(o_t^e) \notag \\
    s_t^p &= \text{MLP}_s([s_t^p, x_t^e]) \in \mathbb{R}^{e} \notag \\
    a_t^p &= \text{Enc}_M(a_t^e)
\end{align}
%At each joint, the position embedding (the index of joint) is considered and concatenated into the latent variable.
 
\noindent \textbf{Casual Transformer} To capture the morphology difference, we apply the prompt technique as in \cite{xu2022PromptDT}, but embedding the morphology specifications instead of imitations
\begin{equation}
    \text{Prompt} = \text{Embed}(K, n, m, x)
\end{equation}
The casual transformer then translates the prompt and the input sequence into the output sequence
\begin{align}
    \text{output} =& \text{Enc}_T(\text{Prompt}, \text{input}) \\
    \text{input} :=& \{\text{BOS}, s_0^p, \quad a_0^p, s_1^p, \quad \cdots, \cdots, \quad a_{t-1}^p, s_t^p  \} \\
    \text{output} :=& \{\hat{s}_0^p, \hat{a}_0^p, \quad \hat{s}_1^p, \hat{a}_2^p, \quad \cdots, \cdots, \quad \hat{s}_t^p, \hat{a}_t^p \}
\end{align}
with a forward casual time mask. Detailed structure is inherited from GPT2, a decoder-only structure as in \cite{chen2021DT, janner2021TT, zheng2022ODT}. For practical consideration, input and output sequences are truncated by a window length $T_w$, with a padding time mask for episodes shorter than $T_w$. The timestep embedding is also considered and concatenated into the latent variable. 

To emphasize the instant impact, we further conduct multi-head attention by querying the target variable and marking the input variable as key and value:
\begin{align}
    \hat{a}_t^p &\leftarrow \hat{a}_t^p + \text{Attention}(\text{Q=}\hat{a}_t^p, \text{K=}s_t^p, \text{V=}s_t^p) \notag \\
    \hat{s}_t^p &\leftarrow \hat{s}_t^p + \text{Attention}(\text{Q=}\hat{s}_t^p, \text{K=}a_{t-1}^p, \text{V=}a_{t-1}^p) \notag
\end{align}

\noindent \textbf{Projector} The task-specific projectors map latent outputs back to the original spaces:
\begin{equation}
    \hat{a}_t = \text{Proj}_a (\hat{a}_t^p),\quad \hat{s}_t = \text{Proj}_s (\hat{s}_t^p), \quad \hat{V}_t = \text{Proj}_V (\hat{s}_t^p) 
\end{equation}
$\text{Embed}_o, \text{Embed}_x, \text{Embed}_a, \text{Embed}_s, \text{Proj}_a, \text{Proj}_s, \text{Proj}_V$ are all modeled as MLPs with LayerNorm and Relu between layers. More configuration details are on the website. % can be found in Appendix.
%More detailed configuration can be found in Table \ref{tab:hyperparameter}.

\subsection{Training paradigm}
ODM has a two-phase training paradigm including pretraining and finetuning, as in Algorithm \ref{alg:odm}.

%In Section \ref{sec:model} we define a uniformed model structure which can be learned by both offline supervised training and on-policy RL. Correspondingly, 
\noindent \textbf{Pretraining} To mimic the learning process of the human infant, we design a curriculum-based learning mechanism in which the training dataset transverses from the easiest to the most complicated one. During each epoch, the current dataset is trained in an auto-regressive manner with two loss terms:
%  \footnote{Including four gym-mujoco environments: hopper, halfcheetah, walker2d and ant, a unity environment: walker and unimal, a mujoco-based environment containing 100 different morphological agents.}
\begin{align}
    L^{\text{imitation}} &= \text{MSE}(\hat{a}_t, a_t^p), \quad L^{\text{prediction}} = \text{MSE}(\hat{s}_t, s_t^p) \notag \\ 
    L^{\text{pretrain}} &= \eta_i L^{\text{imitation}} + \eta_p L^{\text{prediction}}
\end{align}
where $\text{MSE}$ denotes the mean-square-error. $L^{\text{imitation}}$ corresponds to the imitation of action from demonstrations, while $L^{\text{prediction}}$ encourages the agent to reconstruct observations in the existence of casual mask \footnote{$L^{\text{prediction}}(t=0)$ is masked out since it is meaningless to predict the very first state which is randomly initiated.}. 
%  \cite{ha2018RecurrentWorldModel, wu2022DayDreamer}
%  between two continuous variables
% as specified in Figure \ref{fig:model}

\noindent \textbf{Finetuning} one extra predict head is activated to predict the state value $\hat{V}_t$; this head as long as the very last prediction head of $\hat{a}_t$ are employed as outputs of actor and critic:
\begin{equation}
    \hat{V}_t \rightarrow V(s_t), \quad \hat{a}_t \rightarrow \pi(s_t)
\end{equation} % with shared bottom parameters.
%  with more details introduced in Section \ref{sec:prelim_methods}
Actor and critic can then be trained by PPO. Keeping some extent of $L^{\text{pretrain}}$ as auxiliary loss, this finetuning becomes a self-supervised (and model-based, in the aspect of state-action jointly learning) RL
\begin{equation}
    L^{\text{finetune}} = \eta_1 L^{\text{PPO}} + \eta_2 L^{\text{pretrain}}
\end{equation}

% \vspace{-2mm}
\begin{algorithm}[h]
\caption{ODM}
\label{alg:odm}
\begin{algorithmic}[1]
\STATE {\textbf{Initialize} $\theta$}

\STATE {\textbf{Pretraining:}}
\STATE \qquad set num of epoch = 0 
\STATE \qquad \textbf{SWITCH} between 6 body shapes : \\ % hopper, halfcheetah, walker2d, ant, walker and unimal
\STATE \qquad \qquad activate the current env-specific module \\ % grad of
\STATE \qquad \qquad freeze grads of other env-specific modules \\
\STATE \qquad \qquad \textbf{REPEAT} learning from varied pioneers \\ %  of different expert levels
\STATE \qquad \qquad \qquad training on $L^{\text{pretrain}}$ of mini-batch % self-regressive 
\STATE \qquad \qquad increment num of epoch
%\STATE \qquad \qquad stop when num of epoch $>$ max of epoch
%\STATE \qquad save the model checkpoint
\STATE {\textbf{Finetuning:}}
\STATE \qquad {load the model in the target environment} 
%\STATE \qquad {Initialize the replay memory buffer $\mathscr{R} = \{ \}$} 
\STATE \qquad \textbf{REPEAT} iterations: \\
%\STATE \qquad \qquad generate an episode $\tau$ following $\pi_{\theta}(a|s)$
%calculate action $a_t = \pi_{\theta}(a|s_t, h_t)$
\STATE \qquad \qquad \textbf{for} each actor \textbf{do}
\STATE \qquad \qquad \qquad run current policy $\pi$ for $T$ steps
\STATE \qquad \qquad \qquad compute advantages $\hat{A}_0, \cdots, \hat{A}_T$ % estimates
\STATE \qquad \qquad \textbf{end for}
\STATE \qquad \qquad update $\theta$ with $L^{\text{finetune}}$ on a mini-batch %  the surrogate
\STATE \qquad \qquad stop when converges

%\STATE \qquad collect reward $\{ r_t \}$ and the next state $s_{t+1}$
%\STATE \qquad \qquad Calculate episodic return $G_t$ for each step
%\STATE \qquad \qquad Store episode $\tau$ in $\mathscr{R}$;

%\STATE \qquad \textbf{if} $\text{size}(\mathscr{R}) < N$: Continue
%\STATE \qquad \qquad Continue
%\STATE \qquad \textbf{end if}

%\STATE \qquad {Sample a mini-batch of $N$ episodes from $\mathscr{R}$}

\end{algorithmic}
\end{algorithm}
%\vspace{-2mm}

\section{Experiments}
\label{sec:experiment}

This section highlights our experiment setup and results. Further details (such as the videos) can be found on our website.

\subsection{Configurations}
\label{sec:experiment_config}

Table \ref{tab:sysdim} introduces the configuration details of our experiments, including the environments, agents, baselines, and demonstration pioneers. For further details of hyper-parameters, please refer to Table \ref{tab:hyperparameter} in the appendix.

\begin{table*}[h!]
\caption{Environment morphology details}
\label{tab:sysdim}
\begin{center}
\begin{tabular}{lllllllll}
\hline
\multicolumn{1}{l}{\bf Environment}  &\multicolumn{1}{l}{\bf $K$}  &\multicolumn{1}{l}{\bf $n$}  &\multicolumn{1}{l}{\bf $m$}  &\multicolumn{1}{l}{\bf $x$}  &\multicolumn{1}{l}{\bf $m_s$}  &\multicolumn{1}{l}{\bf $m_a$}  &\multicolumn{1}{l}{\bf $n_s$}  &\multicolumn{1}{l}{\bf $n_a$} \\
\hline
swimmer         & 2 & 2 & 1 & 4 & & & 8 & 2 \\
reacher         & 2 & 3 & 1 & 5 & & & 11 & 2\\
hopper          & 3 & 2 & 1 & 5 & & & 11 & 3 \\
halfCheetah     & 6 & 2 & 1 & 5 & & & 17 & 6 \\
walker2D        & 6 & 2 & 1 & 5 & & & 17 & 3 \\
ant             & 8 & 2 & 1 & 95 & & & 111 & 8 \\
humanoid        & 9 & 6 & 3 & 342 & 20 & 10 & 376 & 17 \\
walker          & 16 & 15 & 4 & 18 & 45 & 25 & 243 & 39 \\
unimal          & 12 & 52 & 2 & 1410 & $\ast$ & $\ast$ & 624 & 24 \\
\hline
\end{tabular}
\end{center}
$\ast$: varied from each agent morphology.
\end{table*}

\begin{table*}[h!]
\caption{Pretraining dataset details}
\label{tab:dataset_detail}
%\scriptsize

\begin{center}
\scalebox{0.9}{
\begin{tabular}{ccccccc}
\hline
\multicolumn{1}{l}{\bf Environment}  &\multicolumn{1}{l}{\bf Source}   &\multicolumn{1}{l}{\bf sampling agent} &\multicolumn{1}{l}{\bf  \# samples}  &\multicolumn{1}{l}{\bf \# episodes}  &\multicolumn{1}{l}{\bf ave ep return}  &\multicolumn{1}{l}{\bf ave ep length} \\ 
\hline
\multirow{ 5}{*}{hopper}    & D4RL & expert & 999,494 & 1,027 & 3511.36$\pm$328.59 & 973.22$\pm$97.84 \\
          & D4RL & medium-expert & 1,999,400 & 3,213 & 2089.88$\pm$1039.96 & 622.28$\pm$263.22 \\
          & D4RL & medium & 999,906 & 2,186 & 1422.06$\pm$378.95 & 457.41$\pm$110.88  \\
          & D4RL & medium-replay & 402,000 & 2,041  & 467.30$\pm$511.03 & 196.96$\pm$195.15  \\
          & D4RL & random & 999,996 & 45,239  & 18.40$\pm$17.45 & 22.10$\pm$11.99 \\
\hline
\multirow{ 5}{*}{halfCheetah}  & D4RL & expert & 1,000,000 & 1,000  & 10656.43$\pm$441.68 & 1000.00$\pm$0.0 \\
          & D4RL & medium-expert & 2,000,000 & 2,000 & 7713.38$\pm$2970.24 & 1000.00$\pm$0.0 \\
          & D4RL & medium & 1,000,000 & 1,000  & 4770.33$\pm$355.75 & 1000.00$\pm$0.0 \\
          & D4RL & medium-replay & 202,000 & 202  & 3093.29$\pm$1680.69 & 1000.00$\pm$0.0 \\
          & D4RL & random & 1,000,000 & 1,000  & -288.80$\pm$80.43 & 1000.00$\pm$0.0 \\
\hline
\multirow{ 5}{*}{walker2D}  & D4RL & expert & 999,214 & 1,000 & 4920.51$\pm$136.39 & 999.21$\pm$24.84 \\
          & D4RL & medium-expert & 1,999,209 & 2,190 & 3796.57$\pm$1312.28 & 912.88$\pm$194.62 \\
          & D4RL & medium & 999,995 & 1,190 & 2852.09$\pm$1095.44 & 840.33$\pm$240.13  \\
          & D4RL & medium-replay & 302,000 & 1,093 & 682.70$\pm$895.96 & 276.30$\pm$263.13\\
          & D4RL & random & 999,997 & 48,907  & 1.87$\pm$5.81 & 20.45$\pm$8.46  \\
\hline          
\multirow{ 5}{*}{ant}       & D4RL & expert & 999,877 & 1,034  & 4620.73$\pm$1409.06 & 967.00$\pm$140.87 \\
          & D4RL & medium-expert & 1,999,823 & 2,236  & 3776.93$\pm$1509.93 & 894.38$\pm$243.20 \\
          & D4RL & medium & 999,946 & 1,202  & 3051.06$\pm$1180.59 & 831.90$\pm$290.71 \\
          & D4RL & medium-replay & 302,000 & 485  & 976.05$\pm$1005.71 & 622.68$\pm$140.87 \\
          & D4RL & random & 999,930 & 5,821  & -58.07$\pm$97.76 & 171.78$\pm$281.25 \\
\hline
walker    & self & ppo & 1,001,861 & 7,928 & 262.74$\pm$281.18 & 126.37$\pm$119.17 \\
\hline
unimal    & self & metamorph & 1,638,400 & 3,710 & 2.52$\pm$5.21 & 410.44$\pm$410.57 \\
\hline
\end{tabular}
}
\end{center}

\end{table*}

\subsubsection{Environments \& Agents}

%\textbf{{Bodies, Environments and Tasks}}: 
%\noindent \textbf{Environments \& Agents} 
We practice with enormous agents, environments and tasks, to validate the general knowledge studied by ODM. These scenes include:

\textit{Body shape}: including swimmer (3-joints, no foot), reacher (1-joint and one-side fixed), hopper (1 foot), halfcheetah (2-foot), walker2d (2-foot), ant (4-foot), and humanoid on the gym-mujoco platform \footnote{\url{www.gymlibrary.dev/environments/mujoco/}}; walker (the agent called ragdoll has a realistic humanoid body) \footnote{\url{github.com/Unity-Technologies/ml-agents}} on the unity platform \cite{juliani2020unity}; and finally unimal, a mujoco-based enviroment which contain 100 different morphological agents \cite{gupta2021unimal}. 
% https://
    
\textit{Environment}: flat terrain (FT), variable terrain (VT) or escaping from obstacles.

\textit{Task}: pure locomotion, standing-up (humanoid), or target reaching (reacher, walker).

%We study our ODM on several environments, including eight gym-mujoco (\url{}) environments: swimmer, reacher, hopper, halfcheetah, walker2d, ant; a unity \cite{juliani2020unity}-based environment: walker, and unimal \cite{gupta2021unimal}, a mujoco-based environment. Unimal is a difficult environment which contains  100 different arbitrary morphological agents; while our other choices of environments also provide plentiful diversity of agent morphology: including single-foot (hopper), two-foot (halfcheetah), four-foot (ant) human-body (walker). 

%The pretrained ODM model is then loaded in one of the above environments and the online experiment starts interactively. To test the generalization ability of ODM, we employ it in another four gym-mujoco environments, including swimmer, reacher, humanoid. During this transfer process, task can also switch from the original locomotion on flat terrain, to varied terrain, or escape from the obstacle; or from the locomotion task to the fix-and reach (reacher), body standup or target reaching. We test the ODM performance in both few-shot and zero-shot studys, comparing with necessary baselines.

\noindent Table \ref{tab:sysdim} exhibit all environments and their the morphological dimensions. Environments are from platforms including gym-mujoco, unimal and unity.

\subsubsection{Baselines}
%\noindent \textbf{Baselines} 
We compare ODM with four baselines, each representing a different learning paradigm:

\textit{Metamorph}: a morphological encoding-based online learning method to learn a universal control policy \cite{gupta2022MetaMorph}.

\textit{DT}: As a state-of-the-art offline learning baseline, we implement the decision transformer with the expert action inference as in \cite{lee2022multigameDT} and deal with continuous space as in \cite{chen2021DT}.  We name it DT in the following sections for abbreviation.

\textit{PPO}: The classical on-policy RL algorithm \cite{schulman2017ppo}. Code is cloned from stable-baseline3 in which PPO is in the actor-critic style.

\textit{Random}: The random control policy by sampling each action from uniform distribution from its bounds. This indicates the performance without any prior knowledge especially for the zero-shot case.

\subsubsection{Demonstrating pioneers}
%\noindent \textbf{Demonstrating pioneers} 
For purpose of pretraining, we collect offline data samples of hopper, halfcheetah, walker2d and ant from D4RL \cite{fu2020d4rl}, as sources of pioneer demonstrations. For these environments, D4RL provides datasets sampled from agents of different skill levels, which corresponds to different learning pioneers in our framework, including \textit{expert}, \textit{medium-expert}, \textit{medium}, \textit{medium-replay} and \textit{random}.
%\begin{itemize}
%    \item \textit{expert}: Teachers with the best skill level. Episode samples have the highest averaged returns and longest length, with small standard deviations.
%    \item \textit{medium-expert}: Records from a mixture of experts and medium players. Episodic performance is the second highest but have large deviations.
%    \item \textit{medium}: Records from medium players. Performance is worse than medium-expert and but have smaller deviations.
%    \item \textit{medium-replay}: Records extracted from the learning experiences of medium players, therefore have worse performance than medium. 
%    \item \textit{random}: Records resulted from random policies. Although it has the worse return, we still believe it is valuable to add it into our training curriculum since one could also learn from failure experiences.
%\end{itemize}
We also train some baseline expert agents and use them to sample offline datasets on walker and unimal. This dataset contains more than 25 million data samples, with detailed statistics shown on our website. Within each curriculum, we also rotate demonstrations from the above pioneers for training, as indicated in Algorithm \ref{alg:odm}.
% Appendix, Table \ref{tab:dataset_detail}

%\subsection{Experiment Results}

\subsection{Pretraining}
\label{sec:pretrain_result}
%\noindent \textbf{Pretraining} 

Model is trained with datasets of hopper, halfcheetah, walker2d, ant, walker and unimal, from the easiest to the most complex. Table \ref{tab:dataset_detail} shows statistics of all dataset used in the pretraining phase. Pretraining is computed distributed on 4 workers, each with 8 gpu, 4000 cpu and 400M memory. For both pretraining and finetuning, the ADAM optimizer is applied with the decay weight of 0.01.

\begin{figure}[h!]
\begin{center}
%\framebox[4.0in]{$\;$}
\includegraphics[width=1.2\linewidth]{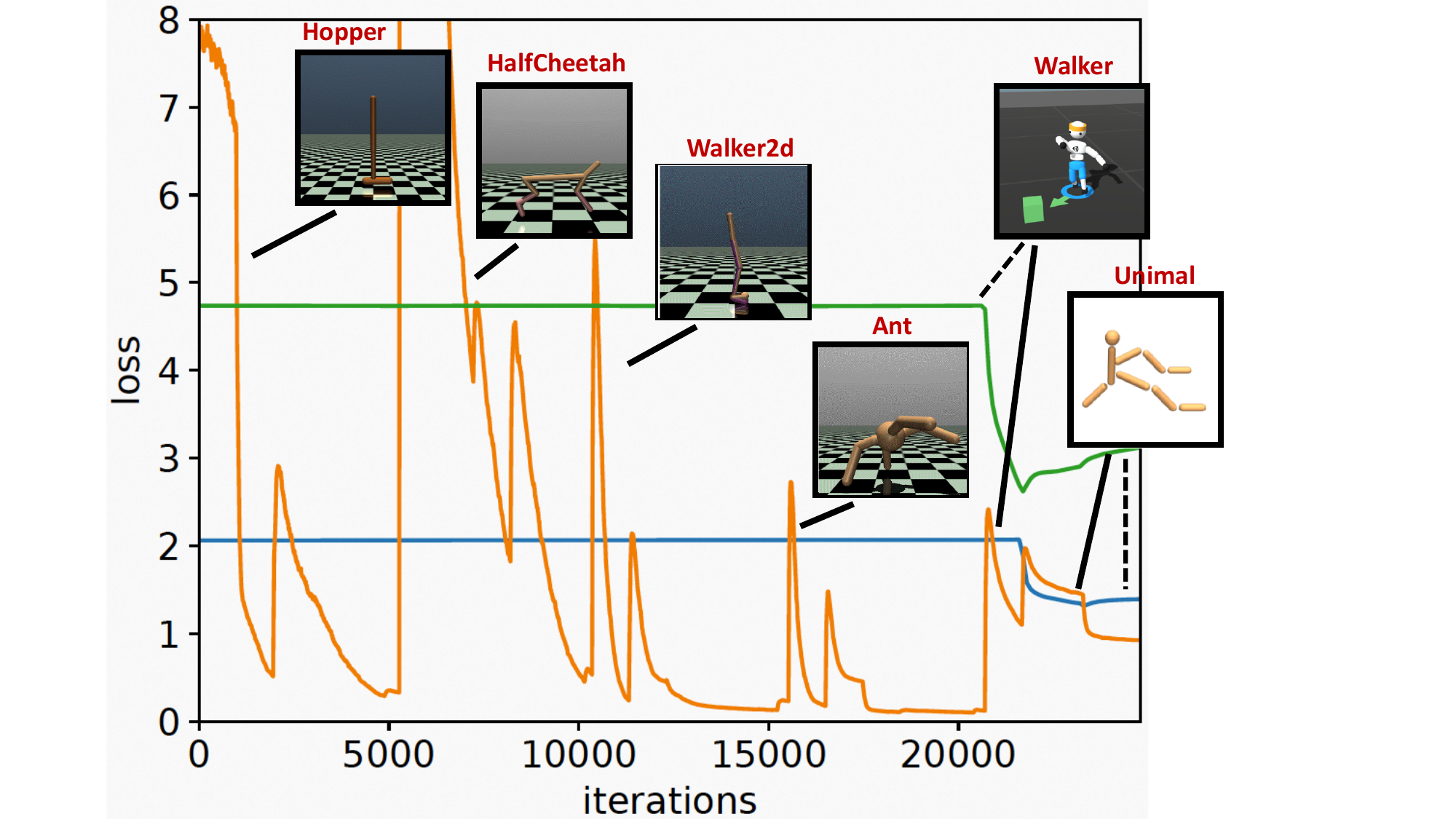} %,height=6cm
%\includegraphics[width=1.0\linewidth]{figure/pretraining_loss.png} %,height=6cm
%\fbox{\rule[-.5cm]{0cm}{4cm}  \rule[-.5cm]{4cm}{0cm}}
\end{center}
\caption{Time plot of pretraining performance. 
Orange: training loss. Green: validation MSE of walker; Blue: validation MSE of unimal. } \label{fig:pretraining}
\end{figure}

Figure \ref{fig:pretraining} shows the loss plot. One can observe that the training loss successfully converges within each curriculum course; although its absolute value occasionally jumps to different levels because of the environment (and the teacher) switching. Validation set accuracy is also improved with walker and unimal as exhibition examples in Figure \ref{fig:pretraining}. 

%\vspace{-2mm}
%\vspace{-1mm}

\begin{figure*}[h!]
\begin{center}
%\framebox[4.0in]{$\;$}
\includegraphics[width=1.0\linewidth]{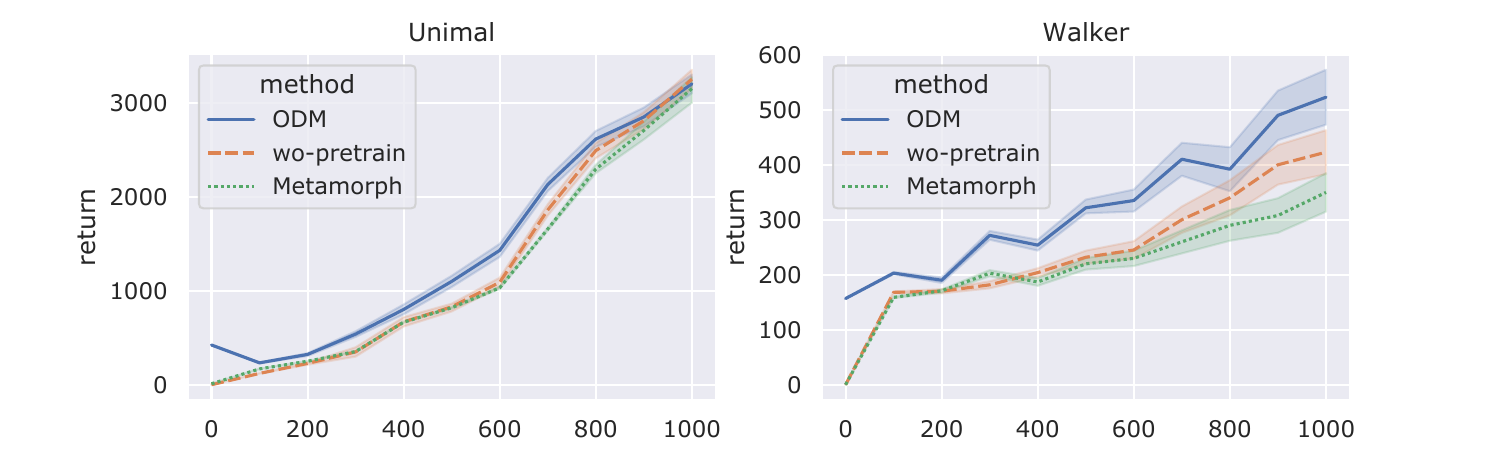} %,height=6cm
%\fbox{\rule[-.5cm]{0cm}{4cm}  \rule[-.5cm]{4cm}{0cm}}
\end{center}
\caption{Comparison of averaged episode returns as functions of iterations during online experiments. Left: unimal. Right: walker. Curves are smoothed and values are rescaled for better visualization purposes.}  \label{fig:onlinez}
\end{figure*}

\begin{table*}[h!]
  \centering
  \caption{Averaged Episodic Performance in online locomotion environments.}
  \label{tab:online-result}
  %\scalebox{0.9}{
  \small
  \begin{tabular}{@{}ccccccc@{}}
    \hline
    \textbf{Metric} & \textbf{Env.} & ODM & MetaMorph & DT & PPO & Random\\
    \hline
    \multirow{ 2}{*}{return} & walker$^\ast$ & \textbf{331.88}$\pm$280.96  & 303.933$\pm$279.16  & 252.74$\pm$281.18 & 265.23$\pm$275.32 & 0.55$\pm$0.83 \\ % [1pt]
                             & unimal & 3197.22$\pm$228.04 & \textbf{3251.68}$\pm$192.61 & -0.09$\pm$3.75    & 2507.32$\pm$260.71 & -3.54$\pm$4.97 \\
    \hline
    \multirow{ 2}{*}{length} & walker$^\ast$ & \textbf{133.29}$\pm$35.88 & 128.42$\pm$33.52 & 126.37$\pm$33.73 & 128.56$\pm$34.06 & 10.34$\pm$1.55\\
                             & unimal & 917.85$\pm$40.84 & \textbf{931.92}$\pm$33.21 & 347.36$\pm$66.47 & 884.39$\pm$50.34 & 321.98$\pm$71.49\\
    \hline
  \end{tabular}\\
 $\ast$: Metrics of walker have substantial deviations since walker has forward process noise implemented.
%}
\end{table*}

\subsection{Online Experiments}
\label{sec:online_result}
%\noindent \textbf{Online Experiments} 
To make online learning faster, we use 32 independent agents to sample the trajectory in parallel, with 1000 as the maximum episode steps. The experiment continues more than 1500 iterations after the performance converges.  Figure 4 provides a quick snapshot of online performances. Compared with ODM without pretraining, returns of ODM are higher at the very beginning, indicating knowledge from pretraining is helpful. As online learning continues, the performance degrades slightly until finally grows up again, and converges faster than the other two methods, although the entire training time (pretraining plus finetuning) is longer. 
% \ref{fig:online}
% , indicating the training criteria of offline supervising and the online RL still have some contradictions
%  evolves for ODM, in comparison with ODM without the pretraining phase (ODM-wo-pretrain), and Metamorph

During online testing, 100 independent episodes are sampled and analyzed to evaluate the agent's performance. The average episode return and episode length are recorded in Table \ref{tab:online-result}. One can observe that our ODM outperforms MetaMorph (in walker) or is similar to MetaMorph (in unimal). Note there are large performance deviations in walker since this environment has substantial process noise implemented to challenge the learning. It is also worth noting that DT does not work for unimal, indicating the limitation of the pure offline method with changing agent body shapes.

\subsection{Few-shot Experiments}
\label{sec:few-shot}

%\noindent \textbf{Few-shot Experiments} 
We examine the transfer ability of policy by providing several few-shot experiments. Pretrained ODM is loaded in several unseen tasks, which are listed in Table \ref{tab:few-shot-result}. As a few-shot test, online training only lasts for 500 steps before testing. ODM obtains the best performance except for humanoid on flat terrain, indicating ODM has better adaptation ability than MetaMorph.

\begin{table*}[h]
  \centering
  \caption{Performance in few-shot experiments.}
  \label{tab:few-shot-result}
  \begin{small}
  \begin{tabular}{@{}ccccccc@{}}
    \hline
    \textbf{Metric} & \textbf{Env.} & Task & ODM & MetaMorph & PPO & Random\\
    \hline
    \multirow{ 6}{*}{return} & unimal & obstacle & \textbf{1611.86}$\pm$179.38  & 1288.15$\pm$127.48 & 932.34$\pm$79.45 & -2.08$\pm$4.85 \\ % [1pt]
                             & unimal & VT & \textbf{580.10}$\pm$41.23 & 499.58$\pm$35.21  & 310.02$\pm$22.93 & -4.87$\pm$8.19 \\
                             & swimmer & FT & \textbf{145.58}$\pm$16.97  & 143.01$\pm$11.31  & 142.36$\pm$13.31 & 0.14$\pm$2.00 \\
                             & reacher & target reaching & \textbf{-32.97}$\pm$5.18 & -33.28$\pm$4.58 & -34.28$\pm$4.56 & -42.96$\pm$0.15 \\
                             & humanoid & FT & 359.90$\pm$54.84 & 360.87$\pm$51.28 & \textbf{360.70}$\pm$50.28 & 108.13$\pm$0.83 \\
                             & humanoid & standup & \textbf{76388.12}$\pm$906.01 & 75750.41$\pm$897.01 & 75033.74$\pm$895.20 & 38921.67$\pm$451.25 \\
    \hline
    \multirow{ 6}{*}{length} & unimal & obstacle & \textbf{827.85}$\pm$53.25 & 771.38$\pm$50.18 & 619.82$\pm$68.35 & 322.05$\pm$65.81 \\
                             & unimal & VT & \textbf{780.23}$\pm$89.02 & 764.48$\pm$77.73 & 524.10$\pm$59.92 & 542.70$\pm$88.76 \\
                             & swimmer & FT & 1000 & 1000 & 1000 & 1000 \\
                             & reacher & reaching & 50$^\ast$ & 50$^\ast$ & 50$^\ast$ & 50$^\ast$ \\
                             & humanoid & FT & \textbf{68.10}$\pm$3.15  & 66.85$\pm$3.94 & 67.59$\pm$2.34 & 22.15$\pm$0.17 \\
                             & humanoid & standup & 1000 & 1000 & 1000 & 1000 \\
    \hline
  \end{tabular}
  
 $\ast$: The official reacher environment has a maximum episode length limit of 50. %We will try to loosen this limit in later studies.
  \end{small}

\end{table*}

%\textbf{1771.70}$\pm$182.34  & 1637.52$\pm$178.60
%\textbf{850.99}$\pm$93.23 & 836.90$\pm$82.16
 
\subsection{Zero-shot Experiments}
\label{sec:zero-shot}
%\noindent \textbf{Zero-shot Experiments} 
Zero-shot experiments can be conducted by inferencing the model directly without any online finetuning. The unimal environment allows such experiment in which the flat terrain (FT) can be replaced by variable terrain (VT) or obstacles. Results are shown in Table \ref{tab:zero-shot-result}. It can be observed that ODM reaches state-of-the-art performance for zero-shot tests, indicating that ODM has strong generalization ability by capturing general high-level knowledge from pretraining, even without any prior experience. 
% In this case, we could still compare ODM with baselines except PPO.

%\vspace{-2mm}
\begin{table*}[h]
  \centering
  \caption{Performance in zero-shot experiments.}
  \label{tab:zero-shot-result}
  \begin{small}
  \begin{tabular}{@{}ccccccc@{}}
    \hline
    \textbf{Metric} & \textbf{Env.} & Task & ODM & MetaMorph & DT & Random\\
    \hline
    \multirow{ 2}{*}{return} & unimal & obstacle & \textbf{1271.70}$\pm$182.34  & 1137.52$\pm$178.60 &  -0.55$\pm$3.06 & -2.08$\pm$4.85 \\ % [1pt]
                             & unimal & VT & \textbf{521.08}$\pm$34.48 & 480.29$\pm$23.21 & 8.22$\pm$6.70 & -4.87$\pm$8.19 \\
    \hline
    \multirow{ 2}{*}{length} & unimal & obstacle & \textbf{750.99}$\pm$86.23 & 736.90$\pm$75.16 & 228.20$\pm$55.74 & 322.05$\pm$65.81 \\
                             & unimal & VT & \textbf{698.80}$\pm$69.49 & 664.63$\pm$72.25 & 585.13$\pm$83.54 & 542.70$\pm$88.76 \\
    \hline
  \end{tabular}
  \end{small}
\end{table*}
%\vspace{-2mm}

\begin{table*}[h!]
  \centering
  \caption{Performance in ablation studies on unimal and walker.}
  \label{tab:ablation-result}
  \scalebox{0.8}{
  \begin{tabular}{@{}cccccccc@{}}
    \hline
    \textbf{Metric} & \textbf{Env.} & Task & ODM & wo pretrain & wo finetune & wo curriculum & wo prompt \\
    \hline
    \multirow{ 4}{*}{return} & unimal & FT & \textbf{3197.22}$\pm$228.04    &   2331.36$\pm$131.24  &  463.12$\pm$84.55  & 434.44$\pm$73.43 & 453.41$\pm$80.13 \\ % [1pt]
                             & unimal & obstacle & \textbf{1611.86}$\pm$179.38    &  592.96$\pm$89.92   &  80.65$\pm$34.91  & 78.34$\pm$32.52 & 75.23$\pm$32.21 \\
                             & unimal & VT & \textbf{580.10}$\pm$41.23    &   404.68$\pm$122.46   &  70.23$\pm$31.31 & 69.34$\pm$30.34 & 70.01$\pm$32.05 \\
                             & walker & FT & \textbf{331.88}$\pm$280.96       &  313.49$\pm$260.35    &    112.64$\pm$72.23 & 109.34$\pm$68.19 & 111.82$\pm$69.76 \\
    \hline
    \multirow{ 4}{*}{length} & unimal & FT & \textbf{917.85}$\pm$40.84     &  845.78$\pm$79.55    &  436.46$\pm$70.0 & 433.24$\pm$70.0 & 436.46$\pm$70.0 \\ % [1pt]
                             & unimal & obstacle & \textbf{827.85}$\pm$53.25     &   554.89$\pm$59.02   &  232.12$\pm$37.02 & 239.12$\pm$29.13 & 230.78$\pm$35.98 \\
                             & unimal & VT & \textbf{780.23}$\pm$89.02     &   526.22$\pm$34.61   &  209.75$\pm$35.41 & 205.51$\pm$33.34 & 204.14$\pm$30.61 \\
                             & walker & FT & \textbf{133.29}$\pm$35.88         &       105.29$\pm$40.13        &      84.32$\pm$34.17 & 80.41$\pm$33.41 & 82.14$\pm$35.43 \\
    \hline
  \end{tabular}
  }
\end{table*}

\subsection{Ablation Studies}
\label{sec:ablation}
%\noindent \textbf{Ablation Studies} 
To verify the effectiveness of each model component, we conduct the ablation tests for ODM with only the online finetuning phase (wo pretrain) and with only pretraining (wo finetune); within the pretraining scope, we further examine ODM without the curriculum mechanism (wo curriculum) and morphology prompt (wo prompt). The DT method could be viewed as the ablation of both $L^{\text{prediction}}$ and $L^{\text{PPO}}$, so we do not list the ablation results of these two loss terms. We conduct the ablation study on unimal (all 3 tasks) as well as walker, with results shown in Table \ref{tab:ablation-result}. Results show that ODM is still the best on all these tasks, which indicating both learning from others' imitation and self-experiences are necessary for intelligence agents.
%  and imitation loss (wo $L^{\text{imitation}}$)

% To verify the effectiveness of the two-phase learning paradigm, we compare with our full version ODM with only one-phase alternative. That is, and with only pretraining (wo finetune). Other ablation analysis including ODM without the morphology prompt (wo prompt)

\subsection{Typical Visualizations}
\label{sec:visualization}
%\noindent \textbf{Typical Visualizations} 
Generalist learning not only aims to improve the mathematical metrics, but also the motion reasonability from human's viewpoint. It is difficult for traditional RL to work on this issue which only solve the mathematical optimization problem. By jointly learning other agent's imitation and bridge with the agent's self-experiences, we assume ODM could obtain more universal intelligence about body control by solving many different types of problems. Here we provide some quick visualizations about generated motions of ODM, comparing with the original versions \footnote{Full version of videos can be found on the website \url{https://rlodm.github.io/odm/}}. 

By examining the agent motion's rationality and smoothness, we first visualize the motions of trained models in the walker environment. Since the walker agent has a humanoid body (the `ragdoll'), the reader could easily evaluate the motion reasonability based on real-life experiences. Figure 5 exhibits key frames of videos at the same time points. In this experiment, we force the agent to start from exactly the same state and remove the process noise. By comparing ODM (the bottom line) with PPO (the upper line), one can see the ODM behaves more like a human while PPO keeps swaying forward and backward and side to side, with unnatural movements such as lowering shoulders and twisting waist.
%  \ref{fig:walker_frames}

\begin{figure}[h!]
\label{fig:walker_frames}
\begin{center}
%\framebox[4.0in]{$\;$}
\includegraphics[width=1.0\linewidth]{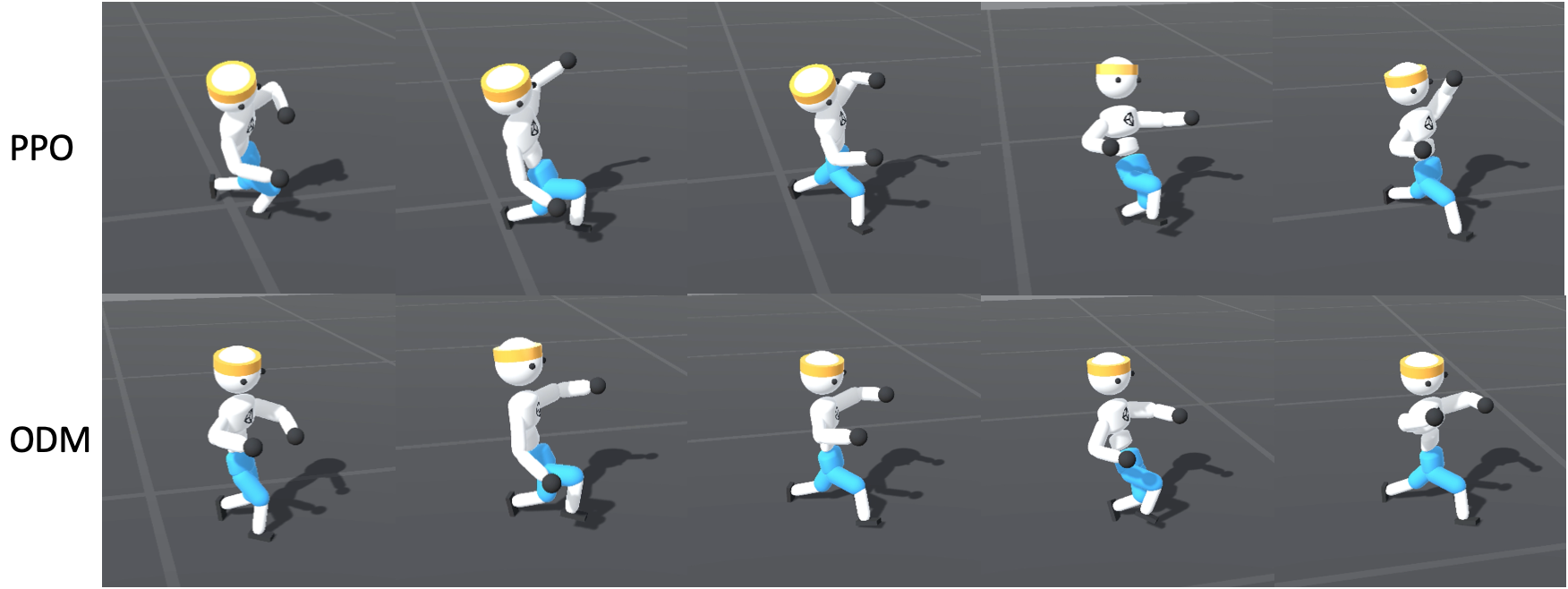} %,height=6cm
%\fbox{\rule[-.5cm]{0cm}{4cm}  \rule[-.5cm]{4cm}{0cm}}
\end{center}
\caption{ODM improves motion fluency and coherence in the walker environment. Keyframes are screened on Second 1, 2, 3, 4, 5, respectively. The video can be found on the website.}
\end{figure}

\begin{figure}[h!]
\label{fig:unimal_frames}
\begin{center}
%\framebox[4.0in]{$\;$}
\includegraphics[width=1.0\linewidth]{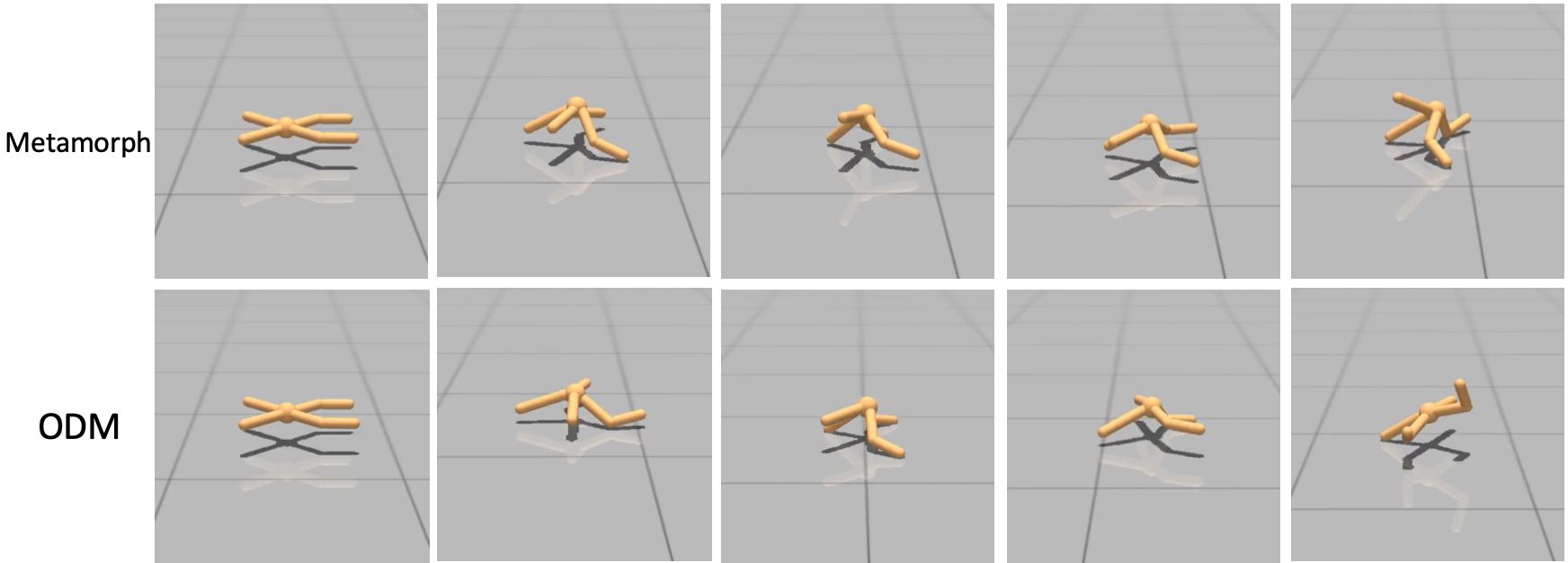} %,height=6cm
%\fbox{\rule[-.5cm]{0cm}{4cm}  \rule[-.5cm]{4cm}{0cm}}
\end{center}
\caption{ODM improves motion agility with visualization of a typical unimal agent. Keyframes are screened evenly from the 30-second video.} % . Video can be found on the website
\end{figure}

We compare the motion agility by visualizing the unimal environment, in which the agent is encouraged to walk toward arbitrary direction. Figure 6 compares ODM with Metamorph. Metamorph wastes most of the time shaking feet, fluid, and gliding, therefore ODM walks a longer distance than Metamorph, within the same time interval \footnote{Figure grids could help the reader to recognize the comparison although the video is more obvious.}.

%  \ref{fig:unimal_frames}

\section{Discussion}
\label{sec:discussion}

Our work can be viewed as an early attempt of an embodied intelligence generalist accommodated for varied body shapes, tasks, and environments. One shortcoming of the current approach is that ODM still has task-specific modules (tokenizers and projectors) for varied body shapes. By using some self-adaptive model structure (e.g. Hypernetwork \cite{ha2017hypernetworks}) in these modules, it is possible to use one unified model to represent the generalist agent. Another potential improvement is to add the value/return prediction into the sequence modeled by the casual transformer. That is, the agent is able to estimate 'the value of its action' before the action is actually conducted, which is also known as 'metacognition' \cite{conwaysmith2022metacognition}. The last interesting topic is the potential training conflict when training switch from offline to online. That might be improved by some hyperparameter tuning (out of this paper's scope), e.g., some warmup schedule of $L^{\text{PPO}}$ during finetuning; but could also be improved by better accommodation of the offline phase knowledge to the online phase. However, these are out of this paper's scope and would require further research and experimentation.

\section{Conclusion}
\label{sec:conclusion}

In this paper, we propose a learning framework to provide a universal body control policy for arbitrary body shapes, environments, and tasks. This work is motivated by the intelligence development process in the natural world, where the agent can learn from others, reinforce with their own experiences, and utilize the world knowledge. To achieve this, we propose a two-dimensional transformer structure that encodes both the state-action morphological information and the time-sequential decision-making process. This framework is designed to be trained in a two-phase training paradigm, where the agent first learns from offline demonstrations of experts on different skill levels and tasks, then interacts with its own environment and reinforces its own skill through on-policy reinforcement learning. By pretraining the demonstration datasets, ODM can quickly warm up and learn the necessary knowledge to perform the desired task, while the target environment continues to reinforce the universal policy. The results of our study suggest that our methodology is able to achieve embodied control over a wide range of tasks and environments, enabling agents to perform complex tasks in real-world scenarios. The results of our study contribute to the study of general artificial intelligence in embodied and cognitive fields. 

\bibliographystyle{plainnat} 
\bibliography{main}

\newpage
\appendix

\section{Additional Methodology Details}
\label{sec:prelim_methods}

During the RL study, the agent generate its action from its policy $\pi(a_t | s_t)$ and push the environment stage forward. This interactive process yields the following episode sequence: 
\begin{equation}
    \tau_{0:T} = \{s_0, a_0, r_0, s_1, a_1, r_1, \cdots, s_T, a_T, r_T\}
\end{equation}
% (\mathcal{S, A, R, P})
in which $T$ means the episode ends or reaching the maximum time length. RL then solves the sequential decision making problem by finding $\pi$ such that $\max_{\pi} \mathbb{E}_{\tau \sim \pi}(R)$ in which the episode return defined as
\begin{equation}
    R := \sum_{t=0}^{T} \gamma^t r_t \label{eq:return}
\end{equation}
with $\gamma \in [0, 1)$ as the discounted factor. 

%Nominal MDP usually consists of the 4-tuple $(\mathcal{S, A, R, P})$, where $\mathcal{S, A, R, P}$ are the state space, action space, set of rewards, and transition probability functions, respectively. 

%Typical online-policy RL methods such as PPO are expected to have higher performances but might subject to higher variances and slower training procedures, comparing with off-policy RL or offline RL. 

The classical PPO methodology inherits from the famous actor-critic framework. The critic generates the state value estimate $V(s)$ , with its loss calculated from Bellman function by bootstrapping the state value function
\begin{equation}
\label{eq:loss_critic}
%\small
L^{\text{Critic}} = \mathbb{E}_{s \sim d^{\pi}}[(r_t + \gamma (V_{\theta_{\text{old}}}(s_{t+1}) - V_{\theta}(s_{t}))^2] 
\notag
\end{equation}
On the other hand, generalized advantage estimation (GAE) \citep{schulman2016gae} is employed to help calculate the action advantage $A_t$  by traversing the episode backward
\begin{align}
    \hat{A}_t &= \delta_t + (\gamma \lambda) \delta_{t+1} + \cdots + (\gamma \lambda)^{T-t+1} \delta_{T-1} \\
    \delta_t &= r_t + \gamma V(s_{t+1}) - V(s_t) \label{eq:gae}
\end{align}

then the actor is learned by maximizing the surrogate objective of $A_t$ according to the policy-gradient theorem
\begin{equation}
\label{eq:loss_actor}
    L^{\text{Actor}} = \text{CLIP}(\text{E}_t \frac{\pi_{\theta_{\text{old}}}(a_t | s_t)}{\pi_{\theta}(a_t | s_t)} \hat{A_t} - \beta \text{KL} [\pi_{\theta_{\text{old}}}(\cdot | s_t), \pi_{\theta}(\cdot | s_t)])
\end{equation}
in which the $\text{CLIP}$ function means clipping the object by $[1 - \epsilon, 1 + \epsilon] \hat{A_t}$, and $\text{KL}$ denotes the famous K-L divergence.

A PPO policy update is then conducted by minimizing the objective upon each iteration:
\begin{equation}
\label{eq:loss_ppo}
L^{\text{PPO}} = -\eta_A L^{\text{Actor}} + \eta_C L^{\text{Critic}}
\notag
\end{equation}

%which augmented the episode to $\tau(\mathcal{S, A, R, P, \hat{A}})$ in the on-policy scenario, with $\mathcal{\hat{A}} = \{ \hat{A}_t \}$.

%Dimensions in Table \label{tab: sysdim}satisfy the relations 
%\begin{equation*}
%    n_s = K*n - m_s + x, \quad n_a = K*m - m_a
%\end{equation*}
%in which $n_s$ and $n_a$ are dimensions of states and actions respectively, according to the conventional RL definitions.

\section{Additional Implementation Details}
\label{sec:hyperparameter}

For practical consideration, input and output sequences are truncated by a window length $T_w$, with padding time mask for episodes shorter than $T_w$. The timestep embedding is also considered and concatenated into the latent variable. 

To emphasize the instant impact, we further conduct a multi-head attention by querying the target variable and marking the input variable as key and value:
\begin{align}
    \hat{a}_t^p &\leftarrow \hat{a}_t^p + \text{Attention}(\text{Q=}\hat{a}_t^p, \text{K=}s_t^p, \text{V=}s_t^p) \notag \\
    \hat{s}_t^p &\leftarrow \hat{s}_t^p + \text{Attention}(\text{Q=}\hat{s}_t^p, \text{K=}a_{t-1}^p, \text{V=}a_{t-1}^p)
\end{align}

Table \ref{tab:hyperparameter} lists most hyper-parameters in our implementation.

\begin{table}[h]
\caption{Hyper-parameters}
\label{tab:hyperparameter}
\begin{center}
\begin{tabular}{lc}
\hline
%\multicolumn{1}{c}{\bf Parameter}  &\multicolumn{1}{c}{\bf Value}
Parameter & Value
\\ \hline
$\gamma$         & 0.90 \\
$\epsilon$    & 0.1 \\
$\beta$    & 0.1 \\
$\lambda$    & 0.1 \\
$N$   & 32 \\
$T$   & 1000 \\
$T_w$   & 10 \\
$e$ & 128 \\
$\eta_A$ & -1.0 \\
$\eta_C$ & 0.1 \\
$\eta_p$ & 0.1 \\
$\eta_i$ & 1.0 \\
$\eta_1$ & 1.0 \\
$\eta_2$ & 0.00001 \\
$\#$ attention head & 2 \\
attention dimension & 1024 \\
$\text{layers}(\text{Embed}_o)$ & $[n, e]$ \\
$\text{layers}(\text{Embed}_x)$ & $[x, e]$ \\
$\text{layers}(\text{Embed}_a)$ & $[m, e]$ \\
$\text{layers}(\text{Embed}_s)$ & $[(K+1)e, e]$ \\
$\text{layers}(\text{Proj}_s)$ & $[e, n_s]$ \\
$\text{layers}(\text{Proj}_a)$ & $[e, n_a]$ \\
$\text{layers}(\text{Proj}_v)$ & $[e, 1]$ \\
learning rate (pretraining) & 0.00001 \\
learning rate (finetuning) & 0.0005 \\
\hline
\end{tabular}
\end{center}
\end{table}

% Within each course, model learn from 5 different levels of pioneers (expert, medium, expert-medium, expert-replay and random). 

\end{document}